\documentclass[10pt,twocolumn,letterpaper]{article}

\usepackage{cvpr}              % To produce the CAMERA-READY version
\usepackage[accsupp]{axessibility}
\definecolor{cvprblue}{rgb}{0.21,0.49,0.74}
\usepackage[pagebackref,breaklinks,colorlinks,allcolors=cvprblue]{hyperref}
\usepackage{algorithm}
\usepackage{multirow}
\usepackage{algpseudocode}
\usepackage{makecell}

\newcommand{\ours}{GRAFT}

\def\paperID{17599} % *** Enter the Paper ID here
\def\confName{CVPR}
\def\confYear{2026}

\title{GRAFT: Graph-Based Affordance Transfer via Part Correspondence
}

\author{
Mengying Lin$^{1}$ \quad
Utkarsh Mishra$^{1}$ \quad
Ajay Mandlekar$^{2}$ \quad
Danfei Xu$^{1}$ \\
\\
$^{1}$Georgia Institute of Technology \quad
$^{2}$NVIDIA \\
\\
{\tt\small \{mlin365, umishra31, danfei\}@gatech.edu} \\
{\tt\small \{amandlekar\}@nvidia.com}
}

\begin{document}
\maketitle
\begin{abstract}
Generalizing robotic manipulation to unseen objects remains challenging, as learning-based approaches require many demonstrations and fail in few-shot settings. Prior work transfers affordances through semantic retrieval, but semantics alone neglect geometric similarity, which is critical for manipulation. We propose GRAFT, a geometry-aware correspondence framework for zero-shot manipulation transfer using only one demonstration per object. Objects are represented as part-based graphs, where part-level descriptors support global instance retrieval and part correspondence, and vertex-level descriptors enable fine-grained contact point matching. For an unseen object, our method first retrieves the most functionally and geometrically similar instance from the demonstration buffer with aligned functional parts, and finally propagates the contact points through point-wise correspondence. 
GRAFT enables zero-shot manipulation transfer through structure-driven correspondence and supports scalable, physically valid demonstration generation via MimicGen. Across zero-shot affordance, physics-based simulation, and real-world evaluations, GRAFT achieves substantially higher correspondence accuracy, manipulation success, and retrieval diversity.

\end{abstract}

\section{Introduction}
\label{sec:intro}
% Realworld exp: (densematcher \& robo-abc)
% Grasp Pan to Grasp Racket
% Open pizza box to Open laptop

Generating diverse robot–object interactions in simulation is an effective way to train manipulation policies that generalize to new objects. A practical strategy for doing so is \emph{retrieval-based data generation}~\cite{ju2024robo-abc, kuang2024ram}: given a target object, the system retrieves one or more related source objects from a dataset together with interactions (e.g., grasps or demonstration trajectories) of how those objects are manipulated. The retrieved interactions serve as templates that can be adapted to new objects, enabling scalable synthesis of realistic training trajectories without manual teleoperation. Crucially, for retrieval to support meaningful transfer, the selected source objects must not only belong to related semantic categories but also share compatible geometric affordances in the specific regions involved in manipulation.

\begin{figure}[t]
    \centering
    \includegraphics[trim={0cm 0.5cm 0.5cm 0},clip,width=\linewidth]{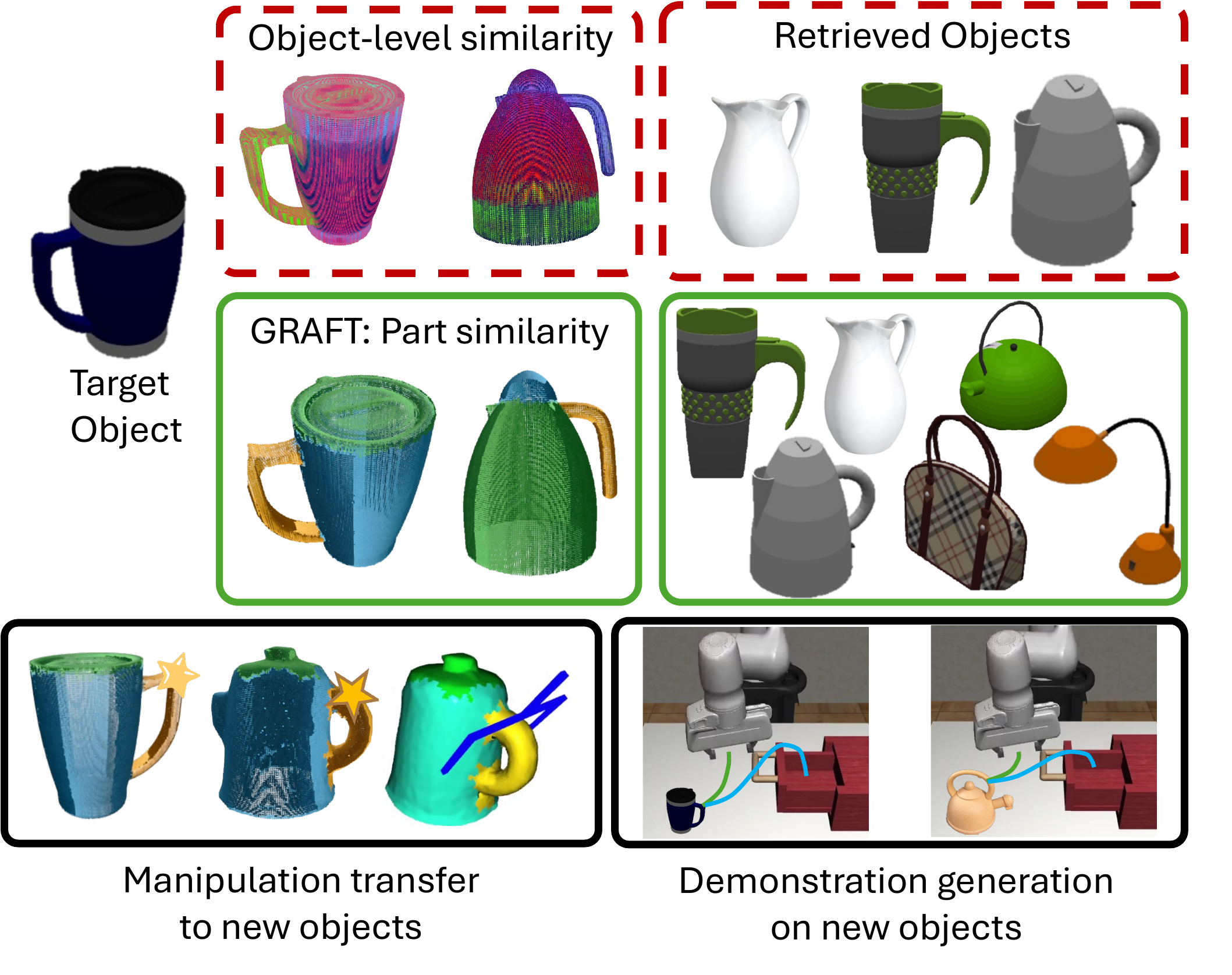}
    \caption{\ours{} enables zero-shot manipulation transfer by reasoning over part-level structural organization, rather than relying on object-level semantic/geometric features. This part similarity approach leads to greater diversity in retrieved objects compared to global similarity pipelines and allows reusing manipulation demonstrations across varied objects with contact point propagation. This capability allows using \ours{} as a manipulation data generator for synthesizing diverse training data for unseen objects from a small set of initial demonstrations on source objects.}
    \label{fig:teaser}
\end{figure}

Existing retrieval pipelines typically follow a two-stage process: first filtering candidates using high-level semantics (e.g., category or textual description)~\cite{ju2024robo-abc, kuang2024ram}, then aligning source and target via geometric correspondence. Recent approaches rely on appearance-based descriptors from foundation models (e.g., DINOv2)~\cite{oquab2023dinov2} as proxies for geometric compatibility, rather than using explicit shape descriptors. The retrieved object serves as a functional analog, what prior work refers to as a digital cousin, enabling zero-shot transfer of contact points or manipulation strategies.
% The retrieved object acts as a functional analog of the target—what prior work refers to as a digital cousin, allowing the demonstrated manipulation to be executed zero-shot on the target by transferring contact points or grasp strategies.

However, we observe that this semantic-to-geometric workflow tends to severely limit object diversity. Semantic filtering over-constrains the candidate pool, while geometric matching is brittle to variations in scale, configuration, and articulation. Small differences in object structure often cause correspondences to fail, which forces retrieval to become conservative and repeatedly select near-identical instances. This reduces the diversity of synthesized interaction data and, in turn, constrains the generalization of policies trained on such data.

Our key insight is that functional compatibility can be determined directly from an object’s \emph{structural organization, rather than holistic semantic features}. Objects that share comparable part configurations often afford similar interactions even if they belong to disjoint categories. This observation motivates a structure-driven retrieval framework that reasons over part-level relationships rather than object-level appearance.

We introduce \underline{Gr}aph-based \underline{A}ffordance \underline{T}ransfer (\ours), a learning-free, geometry-aware correspondence framework that retrieves and aligns objects through their part graph topology. Each object is represented as a graph whose nodes encode multi-level geometric and functional descriptors of individual parts, and whose edges capture spatial relationships between parts. To compare these graphs, we use the Unbalanced Fused Gromov–Wasserstein (UFGW) distance, which jointly measures node-level feature similarity and graph-level structural consistency while allowing partial matching between unequal part sets. This enables robust correspondence even when objects differ in articulation, scale, or number of components.

This capability enables two major outcomes. First, it supports zero-shot transfer of grasps and contact points to entirely new object instances by propagating interaction points through multi-scale correspondences. Second, when integrated with MimicGen~\cite{mandlekar2023mimicgen}, an existing demonstration synthesis system that expands trajectories through motion and scene retargeting, our approach enables the generation of diverse manipulation demonstrations across many object instances from only a small number of seeds. This results in scalable and robust training data generation for manipulation policies that generalize more widely in both simulation and the real world.
\section{Related Work}
\label{sec:related-work}

\textbf{Geometric correspondence and semantic matching.} 
Generalizing manipulation requires dense, semantic-geometric correspondence. Classical methods focused on global shape descriptors or keypoint-based signatures \cite{sun2009hks, bronstein2010si-hks, aubry2011wks}. While effective for rigid alignment, these geometric-only methods fail to capture the functional variability necessary for transfer. Learning-based representations \cite{florence2018dense_object_net, deng2021dif, simeonov2022ndf} enabled contact transfer, a concept later extended by Vision Foundation Model (VFM) \cite{radford2021clip, oquab2023dinov2, tang2023dift}-based methods \cite{ju2024robo-abc, zhu2024densematcher, wang2023d3field, dai2024acdc}, which fuse geometric and semantic cues. 
However, this dependence on object-level geometry or appearance results in brittle correspondence that frequently fails across functional analogs with varied scales or configurations, forcing retrieval to be conservative — objects must look alike to be considered functionally related. Our approach provides a sharp contrast by establishing part-level structural correspondence, aligning objects based on their inherent topological and functional substructures relevant for manipulation, promoting transfer of existing demonstrations grounded purely in functional compatibility.

\textbf{Utilizing correspondence for scalable data generation and policy generalization.} Robust correspondence underpins scalable data synthesis and generalizable policy learning. Affordance-driven frameworks \cite{mo2021where2act, ju2024robo-abc, kuang2024ram, dai2024acdc} use hierarchical semantic-geometric retrieval for knowledge transfer. Furthermore, scalable data generation frameworks \cite{mandlekar2023mimicgen, garrett2024skillmimicgen, jiang2025dexmimicgen} rely on object-centric grasping contact correspondence to retarget demonstrations across new scenes. Yet, their feature-based dependence severely limits dataset diversity, often confining synthesis to near-identical instances. Recent policy methods \cite{wang2024gendp, cheng2023nodtamp} integrate these learned descriptors directly to enhance generalization, but they fundamentally inherit the same issues from feature-based alignment. In contrast, our learning-free, hierarchical alignment, grounded in structural topology, offers a fundamentally more robust and less conservative correspondence foundation, critically boosting data synthesis fidelity and expanding policy generalization capacity.

% traditional HKS, WKS,
% learning baed correspondence,
% correpsondence from fundational models
% \subsection{Generalizable robot manipulation}
% cross-instance, cross-category

\section{Preliminary}

% \textbf{Problem formulation.} 
% \begin{enumerate}
%     \item one target object 
%     \item many source objects with an interaction point (for manipulation)
%     \item we assume given part annotations and bounding box for each part
%     \item the goal is to retrieve similar objects from the source space that can "afford" similar "interactability".
% \end{enumerate}

% \textbf{UFGW full form for graph matching.} 
% \begin{enumerate}
%     \item given target graph from a prescribed target space and a source space, we want to find the closest source graphs from the source space
%     \item UFGW formulates this as a transport problem where .....
%     \item Explain the math that what's going with the UFGW optimization and how do you get the pair-wise scores (target, source candidate)
% \end{enumerate}

\subsection{Problem formulation}
\label{sec:problem-formulation}
We consider the problem of generalizing robotic manipulation to previously unseen object categories. We assume access to a dataset of manipulation demonstrations $\mathcal{D} = \{(o_i, c_i)\}$, where each object instance $o_i$ is associated with a single successful interaction contact $c_i$ for grasping. Given a novel target object $o_{target}$, the challenge is to retrieve the most relevant interaction $(o_j, c_j)$ from $\mathcal{D}$ to guide its manipulation. We assume that all objects, both in the source dataset $\mathcal{D}$ and the target $o_{target}$, have part-level information—encompassing geometric properties for each constituent part (e.g., part-centric point clouds and bounding boxes). A key challenge in this retrieval is that relying on whole-object similarity, which matches objects based on global semantics or geometry, severely limits the diversity of transferable demos. For instance, a "grasp handle" demonstration on a watering can would not be retrieved for a novel mug, despite their shared part-level geometry. Our central hypothesis is that an effective generalization algorithm must instead operate at the part-level to identify functionally similar, yet categorically dissimilar, source objects.

\subsection{Unbalanced Fused Gromov-Wasserstein Distance for Graph Matching}
\label{sec:ufgw}

% A practical approach to process a part-annotated object is to build a graphical representation where each part serves as a node and an edge exists between two parts if the distance between their bounding boxes is below a certain threshold. Then we can formulate the complete problem as a graph matching problem based on some similarity metric. To perform robust matching between our object part graphs, our framework leverages the Unbalanced Fused Gromov-Wasserstein (UFGW)~\cite{titouan2019FGW, sejourne2021UGW} distance as the similarity metric. UFGW is designed to find correspondences, or ``transport plans," between graphs that may have different sizes and structures, making it ideal for our task with the following benefits: (1) directly compares node features (like part geometry), (2) the ``Gromov-Wasserstein" component compares the internal graph structure (the spatial relationships between parts), and (3) the ``unbalanced" nature allows for partial matching, which is crucial for matching a specific part (e.g., a mug handle) to a subset of parts on a different object (e.g., a watering can).
We represent a part-annotated object as a graph, where nodes correspond to parts and edges connect nearby parts based on bounding box proximity. This enables formulation as a graph matching problem. To robustly match part graphs, we use the Unbalanced Fused Gromov–Wasserstein (UFGW) distance~\cite{titouan2019FGW, sejourne2021UGW}, which handles graphs with varying sizes and structures. UFGW jointly compares node features (e.g., geometry) and relational structure, while its unbalanced formulation supports partial matching, allowing specific parts (e.g., a mug handle) to align with subsets of parts on different objects.

Given a set of $N$ source graphs in a dataset,
$$\mathcal{G}_s = (X_s, A_s, \mu_s), \quad s \in \{1,\dots,N\},$$
where:
(1) $X_s = \{x^s_i\}_{i=1}^{n_s} \subset \mathbb{R}^{d_x}$ are node features,
(2) $A_s \in \mathbb{R}^{n_s \times n_s}$ encodes graph structure, including relative translation and rotation between two nodes, and
(3) $\mu_s \in \mathbb{R}^{n_s}_+$ is a nonnegative node mass vector.

Provided a target graph $\mathcal{G}_t = (Y, B, \nu),$
with $Y = \{y_j\}_{j=1}^{m} \subset \mathbb{R}^{d_y}$, structural matrix $B \in \mathbb{R}^{m \times m}$, and node masses $\nu \in \mathbb{R}^m_+$.

For each source $s$, we define a feature cost matrix
$$M^{(s)} \in \mathbb{R}^{n_s \times m}, 
\qquad M^{(s)}_{ij} = c_{\mathrm{feat}}(x^s_i, y_j),$$
where $c_{\mathrm{feat}}$ is a dissimilarity function. Our goal is to find  a transport plan
$T^{(s)} \in \mathbb{R}^{n_s \times m}_+.$
for a fixed source graph $s$, 
% For a fixed source graph $s$, we aim to find a transport plan
% $$T^{(s)} \in \mathbb{R}^{n_s \times m}_+.$$
that approximately maps nodes in the source to nodes in the target. Its row and column sums,
$$u^{(s)} = T^{(s)} \mathbf{1}_m \in \mathbb{R}^{n_s}_+, 
\qquad v^{(s)} = {T^{(s)}}^\top \mathbf{1}_{n_s} \in \mathbb{R}^m_+.$$
represent the transported mass of each node in the source and target, respectively.
The UFGW objective combines structural and feature alignment. The structural discrepancy is
\begin{equation}
    \mathcal{E}_{\mathrm{GW}}(T^{(s)}) 
= \frac{1}{2}\sum_{i,i'=1}^{n_s}\sum_{j,j'=1}^{m} 
\ell_2\!\left( A_s(i,i'), B(j,j') \right) \, T^{(s)}_{ij} \, T^{(s)}_{i'j'},
\end{equation}
where $\ell_2$ denotes the $\ell_2$-norm, measuring differences between structural relations of node pairs. The feature discrepancy is
$$\mathcal{E}_{\mathrm{fuse}}(T^{(s)})
= \sum_{i=1}^{n_s}\sum_{j=1}^{m} M^{(s)}_{ij} \, T^{(s)}_{ij}.$$

The final similarity metric is obtained by solving:
\begin{equation}
\label{eq:ufgw}
    \begin{aligned}
\mathrm{UFGW}_\alpha^\tau&(\mathcal{G}_s, \mathcal{G}_t) 
= \\
\min_{T \in \mathbb{R}^{n_s \times m}_+} 
&\; (1-\alpha)\,\mathcal{E}_{\mathrm{GW}}(T) 
+ \alpha\,\mathcal{E}_{\mathrm{fuse}}(T) \\
&+ \tau_s \, \mathrm{D}_{\mathrm{KL}}\!\left( T\mathbf{1}_m \, \big\| \, \mu_s \right)
+ \tau_t \, \mathrm{D}_{\mathrm{KL}}\!\left( T^\top \mathbf{1}_{n_s} \, \big\| \, \nu \right) \\
&- \varepsilon \, H(T),
\end{aligned}
\end{equation}

where $H(T) = -\sum_{i,j} T_{ij} \log T_{ij}$ is the entropy, $\alpha \in [0,1]$ balances structural vs. feature costs, $\tau_s, \tau_t > 0$ control mass penalization, and $\varepsilon \ge 0$ is an entropic regularizer.

\section{Method}

\begin{figure*}[t]
    \centering
    \includegraphics[trim={0cm 0cm 1cm 0},clip,width=0.9\linewidth]{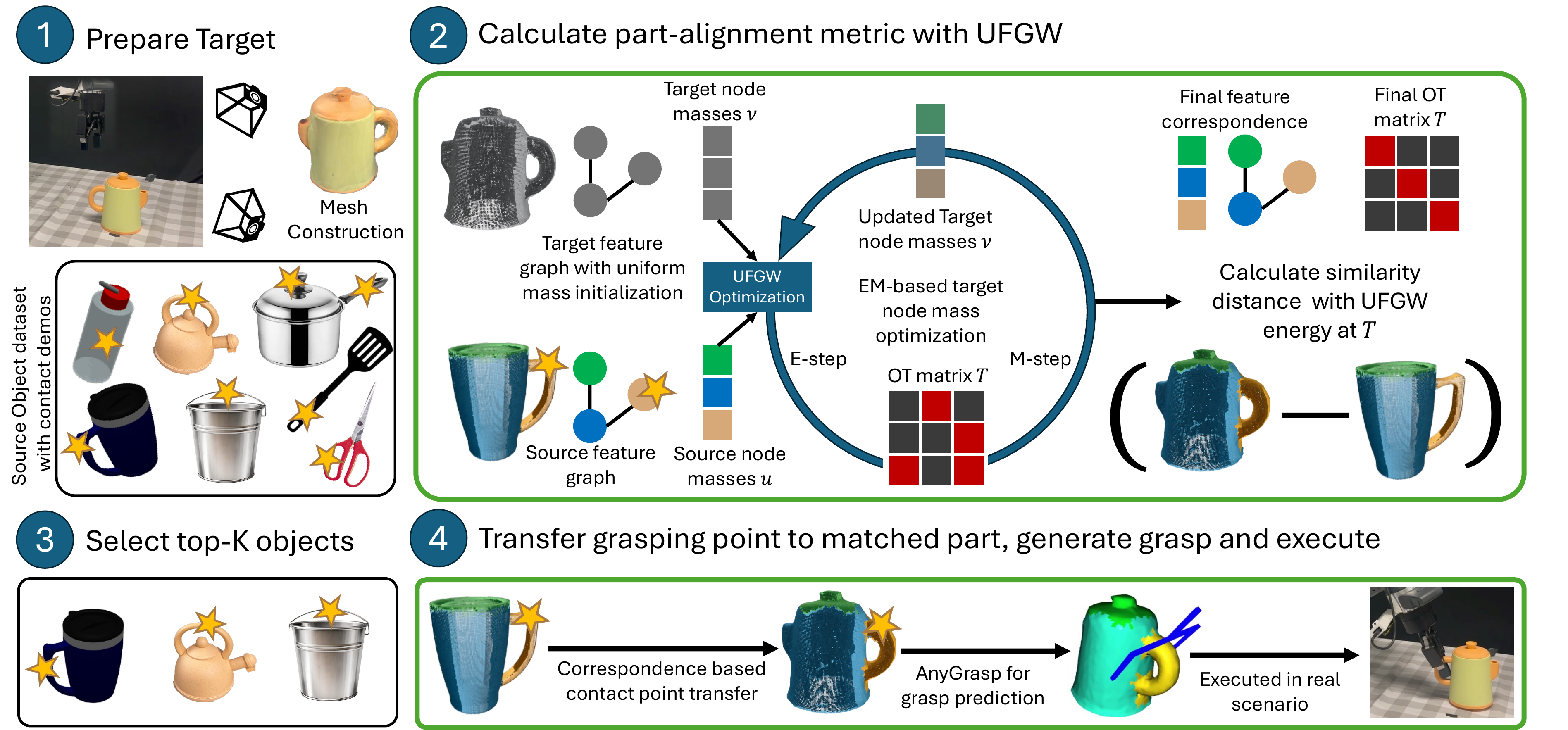}
\caption{
Overview of our correspondence-driven manipulation pipeline. 
\textbf{(1)} Construct a part-graph for the target object from its reconstructed mesh. 
\textbf{(2)} Compute target–source structural correspondence using UFGW with EM-based optimization, with E steps for transport plan solving and M steps for target mass updates. 
\textbf{(3)} Retrieve the top-$K$ most functionally compatible source objects based on the final UFGW energy. 
\textbf{(4)} Propagate contact points through part correspondences to produce zero-shot grasp predictions with AnyGrasp~\cite{fang2023anygrasp} and execute the manipulation in simulation or the real world.
}

    \label{fig:method}
\end{figure*}

% \textbf{Summary.} In summary, we ...

% \subsection{Problem Setup}

% Given a trajectory buffer that contains exactly one human demonstration per object instance in the training set. Each demonstration specifies the contact points and motion trajectory required to manipulate the corresponding object. At test time, the robot encounters a novel object from an unseen category and must produce a feasible manipulation trajectory without collecting additional demonstrations or retraining.

% % Direct imitation learning approaches are ill-suited for this setting, as they require large numbers of demonstrations and tend to memorize category-specific visual patterns rather than generalize across structural variations. Prior retrieval-based methods, such as those relying on semantic similarity, fail to capture the geometric structure that is critical for manipulation tasks. 
% Our key insight is that generalization can be achieved through correspondence when two objects share geometric similarity and functional roles. Thus, if the robot can (i) retrieve a demonstration from a geometrically similar object in the buffer and (ii) establish correspondences between the retrieved object and the target, the contact points and trajectories can be transferred effectively.

We address the problem of retrieving demonstrations from a dataset that can be reliably transferred to previously unseen objects. Prior methods rely on global semantic or geometric similarity, which limits transfer to narrowly related categories. Instead, we focus on part-level structure as the basis for functional compatibility, the property most critical for generalization. Objects that share comparable configurations of functional parts, even when they differ semantically, often support similar interactions. To capture this relationship, we represent each object as an undirected part graph whose nodes encode geometric and functional descriptors of individual parts, and whose edges describe their spatial and relational organization. This representation allows the Unbalanced Fused Gromov–Wasserstein~(UFGW) (\autoref{sec:ufgw}) distance to measure graph similarity and establish reliable correspondences across objects with different shapes, scales, or articulations.

For this framework to be effective, each constructed graph must preserve both the topology and geometry of its constituent parts so that UFGW can infer correspondences purely from structure. This capability forms the foundation of \ours{}, a learning-free hierarchical matching method for robust, multi-scale correspondence. It operates in two connected stages: (1) part-level alignment, which finds fine-grained correspondences for object retrieval, and (2) point-level matching, which localizes transferable contact points. Together, these stages enable zero-shot generalization of grasps and motion trajectories to novel objects.

% To enable the quantification of object functionality similarity, we design a graph-based hierarchical matching framework that operates at three scales:
% \begin{enumerate}
%     \item \textbf{Instance level:} retrieve the most geometrically similar object from the demonstration buffer;
%     \item \textbf{Part level:} identify and align functionally relevant parts between the retrieved object and the target;
%     \item \textbf{Point level:} establish fine-grained correspondences to localize transferable contact points.
% \end{enumerate}
% The retrieved and propagated contact points are then passed to a motion planner (e.g., AnyGrasp) to generate executable manipulation trajectories. Figure~\ref{fig:pipeline} provides an overview of the proposed pipeline.

\subsection{Object feature graph construction}
\label{sec:method-graph-construct}

% We leverage part-annotations for all the objects to enable correspondence reasoning. we represent each object as a structured graph where nodes correspond to object parts and edges capture their spatial relationships. 
% \todo{[what if obj is constructed from real world?] In terms of part annoation, we leverage the fact that many object models (e.g., MJCF or URDF descriptions) already provide explicit structural definitions, including joints, links, and connectivity. This allows us to directly extract a part hierarchy that is both semantically meaningful and physically consistent.}
% Instead of performing unsupervised part decomposition from raw geometry, we leverage the fact that many object models (e.g., MJCF or URDF descriptions) already provide explicit structural definitions, including joints, links, and connectivity. This allows us to directly extract a part hierarchy that is both semantically meaningful and physically consistent.

In order to ensure geometric-affordance aware similarity matching, we construct feature graph representations of all source and target objects. In this graph, nodes encode multi-level descriptors of individual parts, while edges represent their spatial and relational organization. This structured representation preserves both local part geometry and global topology, allowing the UFGW metric (\autoref{sec:ufgw}) to compute functional similarity and establish reliable cross-object correspondences.

\paragraph{Node features.}  
Each node represents an object part and stores multiple levels of descriptors. At the geometric level, we record oriented bounding box (OBB) extents $e \in \mathbb{R}^3$ along with normalized part point clouds $\mathcal{P} = \{p_i \in \mathbb{R}^3\}_{i=1}^N$. To enable point-wise matching within parts, each node also includes vertex-wise descriptors in the form of canonical positional encodings. Specifically, after aligning the part point cloud to its local canonical frame and scaling to $[-1,1]^3$, each vertex $p_i$ is mapped to an encoding
\begin{equation}
    \phi(p_i) = \big[ \sin(2^k \pi p_i), \; \cos(2^k \pi p_i) \big]_{k=0}^{K-1},
    \label{eq:positional-encoding}
\end{equation}
yielding high-frequency positional descriptors $\phi(p_i) \in \mathbb{R}^{6K}$. We set K to 6 in our setup. This Fourier-style encoding follows prior work \cite{mildenhall2021nerf} to lift input coordinates to a higher dimension for richer geometric cues. To summarize global part appearance and provide a coarse functional signature, we include a descriptor $f \in \mathbb{R}^d$ extracted from a foundation model such as DINOv2~\cite{oquab2023dinov2}. Finally, each node carries a functional annotation $y \in \{0,1\}$ indicating whether human demonstrations make contact with the part, thereby identifying interactable regions. The dissimilarity between nodes ($x,y$) from source and target object graphs, respectively, can then be measured using:
\begin{equation}
    c_{feat}(x, y) = \lambda_{\text{obb}} \,  \text{sigmoid}(\|e_{x} - e_y  \|_2) 
+ \lambda_{\text{global}} \,  cos(f_x, f_y)
,
\label{eq:part-cost-fn}
\end{equation}
where $\lambda_{\text{obb}}$ and $\lambda_{\text{global}}$ are constant weights.

\paragraph{Edge features.}  
Edges are derived from the joint structure between parts. Each edge stores the relative transformation between connected parts, including translation and orientation. This captures the assembly structure of the object and constrains feasible correspondences when matching across objects. The edge length between nodes $i$ and $j$ is defined as
\begin{equation}
    l_{ij} \;=\;
    \displaystyle 
    \frac{\|c_i - c_j\|_2 \;+\; \frac{1}{\pi}\angle(a_i,a_j)}
         {1 + \|c_i - c_j\|_2 \;+\; \frac{1}{\pi}\angle(a_i,a_j)}, 
\end{equation}
Here, $c_i \in \mathbb{R}^3$ denotes the center of node $i$, $a_i \in \mathbb{R}^3$ the axis of node $i$, and $\angle(a_i,a_j)$ the angle between the two axes. The first term encodes both relative translation and orientation in a bounded range $(0,1)$.

% \paragraph{Graph representation.}  
% The resulting graph compactly encodes both the geometry and topology of an object. By combining node-level descriptors with connectivity information, it supports similarity evaluation at multiple scales: global instance retrieval, part-level alignment, and point-level contact matching.

% \begin{figure}[h!]
%     \centering
%     \includegraphics[width=0.09\textwidth]{images/obj-full-graph.png}
%     \includegraphics[width=0.09\textwidth]{images/obj-part1.png}
%     \includegraphics[width=0.09\textwidth]{images/obj-part2.png}
%     \includegraphics[width=0.09\textwidth]{images/obj-part3.png}
%     \includegraphics[width=0.09\textwidth]{images/obj-box-graph.png}
%     \caption{An example of the constructed graph.}
% \end{figure}

\begin{figure}[h]
    \centering
    \includegraphics[trim={0cm 1.5cm 0cm 2cm},clip,width=\linewidth]{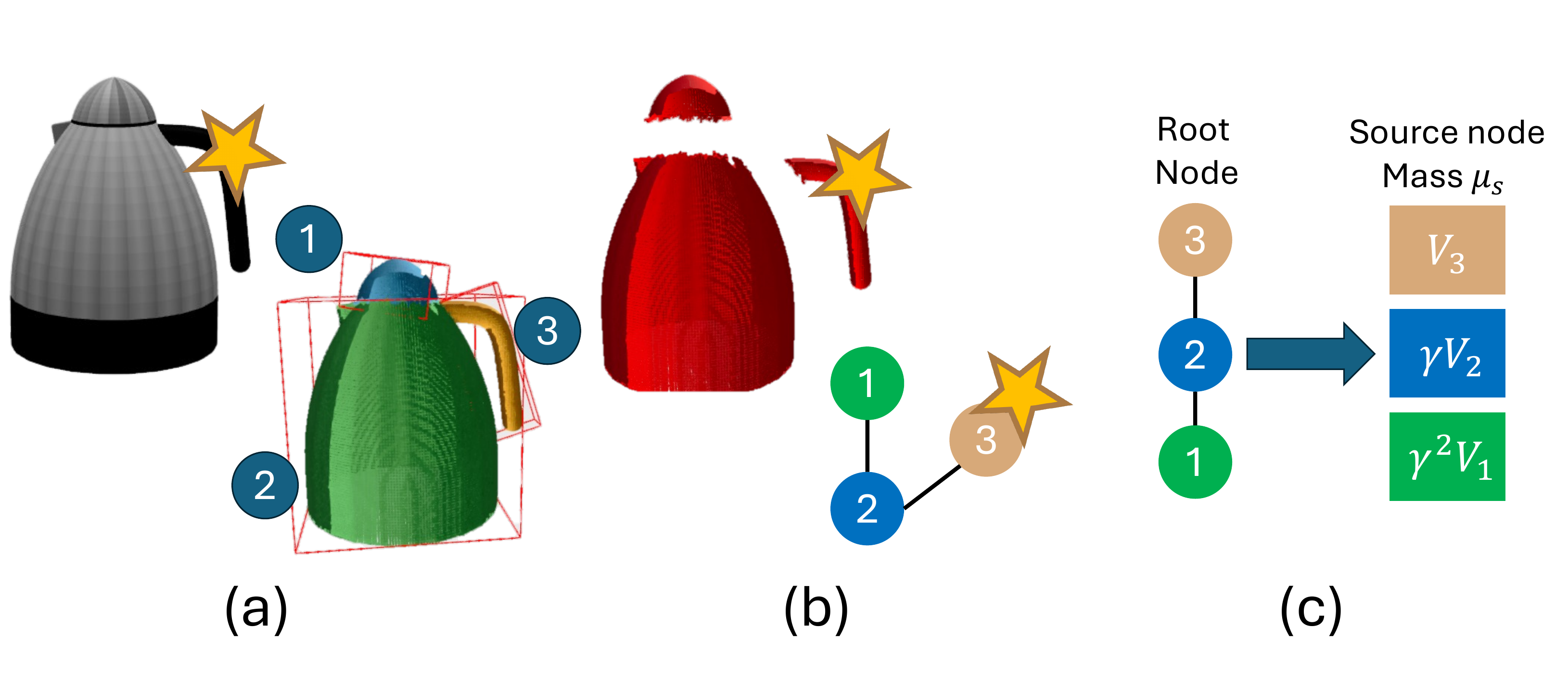}
    \caption{Overview of graph construction and source node mass calculation: \textbf{(a)} The object is decomposed into semantic parts, each represented by an oriented bounding box (OBB). \textbf{(b)} These parts are organized into a graph whose root node corresponds to the part containing the demonstrated grasp contact, enabling functional structuring of the object. \textbf{(c)} Source-node masses $\mu_s$ are obtained as saliency-weighted volumes, where saliency decays exponentially with BFS distance from the root.}
    \label{fig:graph_construction}
\end{figure}

% \subsection{Graph-based Multi-Scale Retrieval and Correspondence}

% Our framework performs retrieval and correspondence in a hierarchical manner. 
% At a high level, we first retrieve the most geometrically similar object from the demonstration buffer, and then propagate contact points to the target object through fine-grained correspondence. 

\subsection{Calculating node mass for object parts}
\label{sec:method-node-mass}

% In the unbalanced optimal transport formulation, each node in a part-graph is assigned a \emph{mass} that encodes its relative importance in the matching process. Intuitively, the mass should capture both the geometric extent of a part and its functional saliency in the context of the interaction. Larger or more salient parts contribute more to the transport plan, while smaller or less relevant parts carry less weight. We distinguish between the source-side masses $\mu_s$ and the target-side masses~$\nu$.

In the UFGW-based matching framework, each node in an object’s feature graph is assigned a mass that determines its relative contribution to the optimal transport objective. Within the unbalanced optimal transport formulation, this mass controls how much influence each part exerts during correspondence estimation. Intuitively, node mass reflects the structural and functional importance of a part—larger or more salient parts should carry higher weight, while smaller or peripheral components should contribute less. We denote source-nodes mass' by $\mu_s$ and target-nodes mass' by $\nu$.

\paragraph{Source node mass.}  
For each source object graph $\mathcal{G}_s$, node masses $\mu_s \in \mathbb{R}^{n_s}_+$ are fixed and derived from both geometric extent and functional saliency. Specifically:
\begin{enumerate}
\item \textbf{Volume-based component:} Each part is represented by its OBB, with its volume as a proxy for size.
\item \textbf{Saliency-based component:} Since geometric size alone does not indicate functional relevance, we propagate saliency through the graph via breadth-first search (BFS) initialized at the interactable part identified in the demonstration as the root node. Saliency decays exponentially with BFS distance, assigning higher weights to nodes closer to the demonstrated contact region. We represent the saliency factor as $\gamma=0.6$. Thus, for the $i^{th}$ source node with an OBB volume $V_i$ and BFS-distance from root node $d_i$, the node mass will be $\mu_{s,i} = \gamma^{d_i}V_i$.
\end{enumerate}
The final mass $\mu_s$ is obtained as the saliency weighted volume, representing both geometric prominence and functional proximity.

\paragraph{Target mass.}  
Unlike the source graph, the target object graph $\mathcal{G}_t$ lacks annotated interaction parts, and thus its node masses $\nu \in \mathbb{R}^m_+$ cannot be predefined. To address this, we infer $\nu$ jointly with the transport plan $T$ through an alternating optimization procedure integrated into the UFGW algorithm. The update follows an expectation–maximization (EM) pattern:
\begin{itemize}
\item \textbf{E-step:} With the current target mass $\nu$ fixed, we solve for the optimal transport plan $T$ by minimizing the UFGW objective~\autoref{eq:ufgw}.
\item \textbf{M-step:} Using the obtained transport plan $T$, compute the empirical incoming flow to target nodes, $\tilde{\nu}\;=\;T^\top \mathbf{1}_{n_s} \in \mathbb{R}^m_+$
  and update the target mass via a convex combination with a geometric prior $\delta_{vol} \in \mathbb{R}^m_+$, calculated as the target parts' normalized OBB volumes:
  \[
    \nu^{\mathrm{new}} \;=\; (1-\lambda)\,\delta_{vol} \;+\; \lambda\,\tilde{\nu},\qquad \lambda\in[0,1],
  \]
\end{itemize}
This update balances prior geometric knowledge ($\lambda=0$) and transport-induced evidence ($\lambda=1$). Empirically, we set $\lambda=0.4$ to stabilize convergence and prevent degenerate concentration of mass on a few target nodes. The resulting masses adaptively weight each part’s participation in the correspondence, improving robustness of the UFGW optimization to variations in part scale and interaction relevance.

% \paragraph{Remarks.}
% \begin{itemize}
%   \item The convex-blend update is simple and effective: the prior $\pi$ injects geometric / heuristic knowledge (e.g.\ volumes or seeded saliency), while the empirical flow $\tilde{\nu}$ lets the solver correct the prior based on the matching.
%   \item The KL penalty parameter $\tau_t$ appearing in the UFGW objective controls how strongly the optimisation penalizes deviation between the empirical flow and the current $\nu$. Large $\tau_t$ forces $\tilde{\nu}$ and $\nu$ to stay close, which reduces the aggressiveness required of the convex blend.
%   \item An alternative principled M-step (when using a quadratic prior or extra penalty terms) can be obtained by minimizing a penalized divergence
%   \[
%     \min_{\nu\ge0}\; \tau_t\,\mathrm{D}_{\mathrm{KL}}(\tilde{\nu}\,\|\,\nu) \;+\; \eta\,\mathcal{R}(\nu,\pi),
%   \]
%   where $\mathcal{R}$ encodes the prior (for example $\mathcal{R}(\nu,\pi)=\lVert \nu-\pi\rVert_2^2$). For simple choices of $\mathcal{R}$ this leads to closed-form or easily computable updates; the convex blend above is the most transparent and computationally cheap choice.
% \end{itemize}

\subsection{\ours: \underline{Gr}aph-based \underline{A}ffordance \underline{T}ransfer}
\label{sec:method-graft}
Our goal is to, given a target object, retrieve from the dataset the most functionally similar object at the part level, and transfer its demonstrated grasp to the target by aligning corresponding parts. To achieve this, \ours{} integrates part-level feature graphs, importance mass estimation for each node via an EM procedure, and UFGW optimization based distance metric into a unified retrieval framework. The optimization takes around 2 ms per
source–target pair, with additional runtime analysis provided in the supplementary material.

\begin{algorithm}[ht]
\caption{\ours: \underline{Gr}aph-based \underline{A}ffordance \underline{T}ransfer}
\label{alg:hierarchical-ufgw}
\begin{algorithmic}[1]
\Require Demonstration dataset $\{(\mathcal G_s,c_s)\}_{s=1}^N$, target $\mathcal G_t$, parameters $\alpha,\tau_s,\tau_t,\varepsilon$, number of elites $K$
% \Output Retrieved index $s^\star$ and part matching plan $T$ on the target
    
\State \textbf{Instance-level scoring:}
\For{$s=1$ \textbf{to} $N$}
    \State $(T^{(s)}, d_s) \gets \textsc{UFGW}(\mathcal G_s,\mathcal G_t,\alpha,\tau_s,\tau_t,\varepsilon)$
\EndFor
\State $\mathcal S_K \gets$ set of best-$K$ tuple $(\mathcal G_s, c_s, d_s)$ with lowest $d_s$ \\

\Return $S_K$
\end{algorithmic}
\end{algorithm}

\begin{algorithm}[t]
\caption{\textsc{UFGW}}
\label{alg:ufgw-persource}
\begin{algorithmic}[1]
\Require Source $\mathcal G_s=(X_s,A_s,\mu_s)$, target $\mathcal G_t=(Y,B,\nu)$, parameters $\alpha,\tau_s,\tau_t,\varepsilon$, tolerances

\State Initialize $T \gets \operatorname{normalize}(\mu_s\,\nu^\top)$
\For{outer = 1 \textbf{to} max\_outer\_iter}
    \State Compute feature cost $M^{(s)}$
    \State Compute structural cost:
    \[
    C_{\mathrm{GW}}(i,j) \gets \sum_{i',j'} \ell_2\big(A_s(i,i'),B(j,j')\big)\;T_{i'j'}
    \]
    \State $T^\star \gets$ \textsc{Solve Optimization \autoref{eq:ufgw}}
    \If{$\|T^\star-T\|_1 / \|T\|_1 < \text{tol}$}
        \State \textbf{break}
    \EndIf
    \State $T \gets T^\star$
\EndFor
\State Compute $d_s$ by evaluating UFGW distance at $T^\star$\\
\Return $T^\star,d_s$
\end{algorithmic}
\end{algorithm}

\textbf{Part-level alignment-based retrieval.} The UFGW formulation jointly considers node features and structural dissimilarities while allowing mass variation through KL-penalized marginal constraints. This facilitates the EM-update to find the optimal node masses for the target graph. Performing the two steps iteratively for each (source object, target object) tuple improves the estimation of the transport plan and the accuracy of the UFGW distance between them. Finally, we select the best-K source objects and their corresponding demos with lowest distance from target object.

% \subsubsection{Instance Retrieval with Unbalanced Fused Gromov-Wasserstein Distance}
% Given a part-annotated test-time object, our goal is to retrieve the source object from the demonstration buffer that exhibits the most similar functionality. To achieve this, we represent each object as a graph with nodes corresponding to parts and edges encoding their geometric or structural relations. We then compute the Unbalanced Fused Gromov-Wasserstein (UFGW) distance between the target and each source graph. The UFGW formulation jointly considers node features and structural dissimilarities while allowing mass variation through KL-penalized marginal constraints. The source graph with the lowest UFGW cost is selected as the closest functional match, and its optimal transport plan provides a soft correspondence between parts.

\textbf{Grasping contact point transfer.} For the selected best-K source objects, we can calculate the feature cost matrix for each of them to identify the part in target object that aligns the most with the interacted part in the source object. 
To achieve robust point-level alignment, we (1) leverage the part-level vertex descriptors defined in \eqref{eq:positional-encoding}, and (2) compute dense correspondences between the key source part and its matched target part. 
This mapping enables direct transfer of grasping contact points from the demonstration to the target object. 
The propagated contact points are finally passed to a grasp planner (e.g., AnyGrasp~\cite{fang2023anygrasp}) and then to a motion planner (e.g. RRT-Planner~\cite{kuffner2000rrt}) to generate executable manipulation trajectories.

\section{Experiments}

In this section, we evaluate \ours{} for zero-shot manipulation, which establishes structural alignment between objects by modeling part-level relations through graph matching. Our central hypothesis is that explicitly reasoning over part connectivity and functional dependencies, rather than relying solely on object-level feature similarity—enables more reliable manipulation transfer to unseen object categories. To test this, we conduct experiments across three settings: (i) zero-shot affordance and execution evaluation to assess manipulation generalization, (ii) data generation via MimicGen~\cite{mandlekar2023mimicgen} to test correspondence-guided trajectory transfer, and (iii) ablation studies to analyze the effect of structural and optimization components.

\subsection{Zero-shot manipulation generalization}
\label{sec:exp-zero-shot}

We evaluate our method for zero-shot manipulation capabilities, providing no demonstrations or fine-tuning for test object categories. While physical task success is the ultimate goal, simulation results conflate multiple sources of error from correspondence (our algorithm) and motion planning (required for evaluating grasp success). To isolate correspondence quality from planning errors, we employ a two-stage evaluation: (i) \textit{affordance generalization} to assess the model’s ability to localize graspable or functionally relevant regions on unseen objects, and (ii) \textit{simulation-based execution}, to test whether the transferred grasp and trajectory remain physically executable under simulated dynamics.

% While the ultimate goal of manipulation is physical task success, evaluating only simulation outcomes conflates multiple sources of error, including correspondence, and motion planning and execution. 
% To disentangle these factors, we adopt a two-stage evaluation protocol. 

% The first stage, \textit{affordance generalization}, measures whether the model can correctly localize graspable or functionally relevant regions on unseen objects—reflecting its understanding of geometric and functional similarity. 
% The second stage, \textit{simulation-based execution}, evaluates whether the transferred grasp and trajectory can be successfully executed under physical dynamics. 
% % This separation allows us to diagnose whether performance gains stem from better structural correspondence or from improved motion feasibility.

\subsubsection{Affordance Generalization}

\paragraph{Setup.} 
We follow the evaluation setup of ROBO-ABC~\cite{ju2024robo-abc}. Each test object is manually annotated in Blender with a binary mask marking graspable regions. Since graspability varies smoothly near contact boundaries, we convolve the mask with a Gaussian kernel to obtain a continuous \textit{affordance score field}, following common practice in affordance evaluation~\cite{,fang2018demo2vec, luo2022AGD20K}.

% However, graspability is not strictly discrete, points near the annotated contact region often remain partially graspable depending on gripper tolerance and local geometry. 
% To capture this spatial continuity, we convolve the binary mask with a Gaussian kernel to obtain a continuous \textit{affordance score field}, which serves as the ground-truth graspability map. 
% This design follows common practice in affordance and visual saliency. evaluation~\cite{ju2024robo-abc,mo2021where2act}, where a smoothed distribution better reflects physical plausibility
% and ensures stable computation of continuous-valued metrics such as normalized scanpath saliency (NSS) and distance-to-mass (DTM). 
% In our implementation, the kernel standard deviation $\sigma$ is adaptively set to $0.02\,d_{\text{bbox}}$, where $d_{\text{bbox}}$ denotes the diagonal length of the object’s bounding box, ensuring consistent spatial scaling across object sizes.

\paragraph{Evaluation.} We extend ROBO-ABC~\cite{ju2024robo-abc} affordance metrics to 3D point clouds, representing each object by ground-truth points ${p_i, w_i}$ with graspability weights. Given predicted grasp points $\mathcal{P}$, we compute: (i) \textbf{ASR}—fraction of predictions in graspable regions, (ii) \textbf{NSS}: average normalized saliency within the affordance field, and (iii) \textbf{DTM}: mean normalized distance from the nearest graspable region. More details about the metrics are provided in the supplementary. 

\begin{table}[h]
\centering
\resizebox{0.7\linewidth}{!}{
\begin{tabular}{lccc}
\toprule
\textbf{Methods} & \textbf{ASR}$\uparrow$ & \textbf{NSS}$\uparrow$ & \textbf{DTM}$\downarrow$ \\
\midrule
where2act~\cite{mo2021where2act}    & 0.31 & -0.13 & 0.14 \\
\makecell{ROBO-ABC~\cite{ju2024robo-abc} \\ (CLIP+DIFT)}      & 0.54 & -0.10 & 0.12 \\
\makecell{ROBO-ABC~\cite{ju2024robo-abc} \\ (DINO+DIFT)}      & 0.69 & 0.48 & 0.10 \\
\midrule
\textbf{\ours} & \textbf{0.85} & \textbf{0.82} & \textbf{0.02}\\
% \textbf{XXX(Ours)} & 0.76 & 1.62 & 0.01 \\
%    
\bottomrule
\end{tabular}
}
\caption{Affordance generalization results comparing our method with prior approaches. 
\ours{} achieves the highest ASR and NSS and the lowest DTM, indicating more accurate localization of graspable regions on unseen objects.}
\label{tab:results}
\end{table}

Compared to prior methods such as where2act~\cite{mo2021where2act} and ROBO-ABC~\cite{ju2024robo-abc}, our approach achieves higher Success Rate (SR) and Normalized Semantic Similarity (NSS), as illustrated in~\autoref{tab:results}, indicating stronger generalization and semantic consistency, while significantly reducing the Distance-to-Motion (DTM), demonstrating more precise correspondence alignment. We also observe that replacing CLIP with DINO in ROBO-ABC improves both SR and NSS, suggesting that DINO’s self-supervised visual features better capture geometric correspondences than CLIP’s text-aligned embeddings, which are less sensitive to fine-grained structural variations.

\subsubsection{Simulation-based Execution Evaluation}

While affordance metrics evaluate perceptual localization, they overlook physical executability.
We therefore conduct physics-based rollouts in SAPIEN~\cite{xiang2020sapien} using a floating gripper, allowing free SE(3) motion without kinematic or planning constraints.
This isolates correspondence and trajectory transfer quality from confounding factors like inverse kinematics (due to object initialization) or motion planning failures (failure in obstacle avoidance), ensuring errors reflect true correspondence or contact inaccuracies.

\paragraph{Evaluation.}
We collect one demo per seen object including \textit{WaterBucket, Teapot, WineBottle, Mug}.
Unseen objects include \textit{Detergent, Shampoo, Spray, Kettle, Briefcase, Handbag, Phone and Remote}.
For each unseen object, we perform 12 randomized trials, spawning the object at different orientations and positions within the workspace and executing the transferred grasp trajectory without adaptation. A trial is considered successful if the gripper achieves a stable grasp and completes the lifting task without slipping, and the overall Simulation Success Rate (SR) is computed as $\text{SR} = N_{\text{success}}/{N_{\text{total}}}$. To further analyze performance, failures are categorized into three types: (1) \textit{AnyGrasp Error} \textbf{(AG)}, where no valid gripper pose is predicted near the target grasp point, indicating a poorly aligned transferred grasp prior; (2) \textit{Grasp Failure} \textbf{(GF)}, where the gripper closes without securing the object, usually due to inaccurate contact localization or misaligned surface normals; and (3) \textit{Slippage} \textbf{(SL)}, where the object is initially grasped but slips or drops during lifting, reflecting partial contact misalignment or insufficient frictional stability. In fact, \textit{AnyGrasp Error} serves as a direct indicator of the grasp transfer and correspondence quality of the baselines.

\begin{table}[t]
\centering
\resizebox{\linewidth}{!}{
\begin{tabular}{lcccc}
\hline
\textbf{Method} & \textbf{SR(\%)}$\uparrow$ & \textbf{AG(\%)}$\downarrow$ & \textbf{GF(\%)}$\downarrow$ & \textbf{SL(\%)}$\downarrow$ \\
\hline
where2act~\cite{mo2021where2act}     
& 36.46 & \textbf{0.00} & 58.33 & \textbf{5.21} \\

AnyGrasp~\cite{fang2023anygrasp}      
& 59.38 & \textbf{0.00} & 19.79 & 20.83 \\

\makecell{ROBO-ABC~\cite{ju2024robo-abc} \\ (CLIP+DIFT)}
& 42.71 & 25.00 & \textbf{12.50} & 19.79 \\

\makecell{ROBO-ABC~\cite{ju2024robo-abc} \\ (DINO+DIFT)}
& 43.75 & 25.00 & \textbf{12.50} & 18.75 \\
\hline
\textbf{\ours}
& \textbf{81.25} & \textbf{0.00} & \textbf{12.50} & 6.25 \\
\hline
\end{tabular}
}
\caption{Simulation Success Rate (SR) and Failure Mode Breakdown in the Floating-Gripper Setup. 
\ours{} achieves the highest SR while substantially reducing grasp failures and slippage, demonstrating more reliable contact transfer and execution.}
\label{tab:sim_results}

\end{table}

We summarize the simulation success rates and failure mode breakdown for different methods in~\autoref{tab:sim_results} in the floating-gripper setup. Our approach achieves the highest overall SR at $81.25\%$, significantly outperforming baselines. Notably, \textit{AnyGrasp Error} is zero for our method, indicating robust and accurate grasp transfer, whereas CLIP- and DINO-based ROBO-ABC variants exhibit high AnyGrasp Error rates, highlighting limitations in correspondence quality. Grasp failures remain the dominant source of error for most methods, reflecting challenges in precise contact alignment, while slippage occurs primarily in methods that succeed in initial grasping but lack stable contact. Overall, these results demonstrate that explicitly reasoning over part-level correspondences substantially improves both transfer reliability and execution feasibility.

\begin{figure}[h]
    \centering
    \includegraphics[trim={0cm 1cm 1cm 0},clip,width=0.8\linewidth]{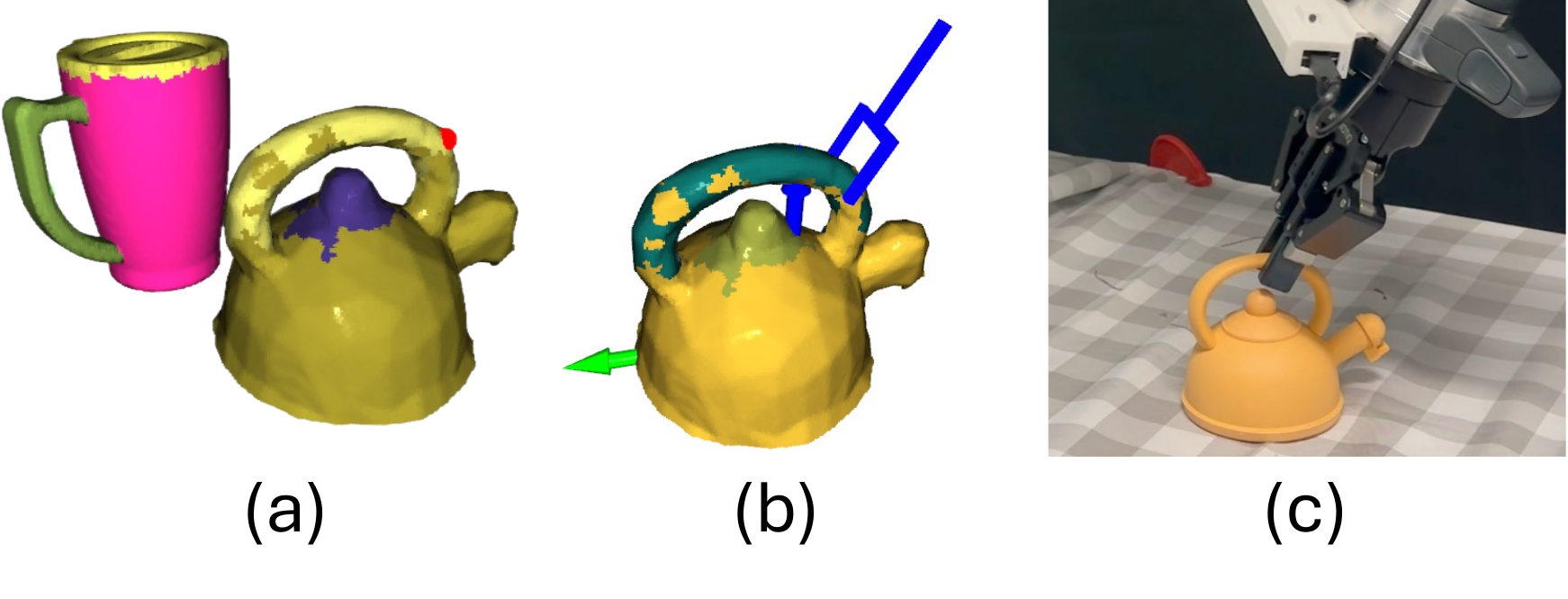}
    \caption{Real-world setup: the target object is first scanned and annotated to obtain object graph. GRAFT then transfers a grasp point from the most compatible source object. A feasible grasp around the transferred point is further generated by AnyGrasp, which is executed on a Franka Panda robot.}
    \label{fig:placeholder}
\end{figure}

\textbf{Real world execution.} We obtain a high-resolution mesh of the target object using a mobile 3D-scanning application~\cite{3dscannerapp} and manually annotate its constituent parts. Using the simulation manipulation dataset, we identify the most compatible source objects using \ours{} and transfer the associated grasping point to the target object. We then employ AnyGrasp~\cite{fang2023anygrasp} to compute a feasible grasp around the grasping point relative to the object’s canonical base frame, and use FoundationPose~\cite{ornek2024foundationpose} to estimate the object’s pose in the real scene with RGBD readings from Azure Kinect camera. The resulting grasp is executed on a Franka Panda robot arm.

\subsection{\ours{} as Data Generator}
\label{sec:exp-data-gen}

Beyond zero-shot execution, our correspondence framework can also serve as a structural prior for large-scale data generation. 
To demonstrate this, we integrate our method with MimicGen~\cite{mandlekar2023mimicgen}, which synthesizes demonstration trajectories by transferring existing ones to new objects.

\paragraph{Trajectory decomposition.}
Given a retrieved source demonstration and the corresponding target object, we divide each transferred trajectory into two stages:
(1)~the \textit{reaching stage}, where the end-effector approaches the contact pose, and 
(2)~the \textit{interaction stage}, where it physically manipulates the object after contact.
The reaching behavior is largely invariant of source demonstrations, since its goal is simply to position the gripper to the predicted contact pose. 
Hence, we regenerate this stage directly toward the predicted grasp point using an RRT planner~\cite{kuffner2000rrt}, without relying on the source trajectory.  
In contrast, the interaction phase depends strongly on object functional alignment, and shall be transferred from source demonstrations.  
We preserve the relative transformations across time steps from the source demonstration, but reorient the entire trajectory according to the predicted grasp pose on the target.  
This ensures that the transferred trajectory maintains the task-relevant motion pattern while adapting to new part geometry.

\paragraph{Data generation setup.}
Using this procedure, we generate a set of synthetic manipulation demonstrations across unseen objects. 
Each demonstration is built by pairing a source demonstration with a structurally similar target object identified via our graph correspondence framework.  
For comparison, we also employ a vanilla MimicGen~\cite{mandlekar2023mimicgen} baseline that transfers trajectories directly between objects without any correspondence reasoning, treating all target objects as geometrically identical to their sources.

\paragraph{Evaluation and results.}
We evaluate both approaches with data generation success rate in Mujoco~\cite{todorov2012mujoco} simulator for 500 generation runs.  
Our method achieves a substantially higher trajectory success rate (63\%) compared with vanilla MimicGen~\cite{mandlekar2023mimicgen} (11\%), demonstrating that correspondence-aware trajectory adaptation significantly improves the physical feasibility of generated data.  

\begin{table}[h!]
\centering
\resizebox{0.7\linewidth}{!}{
\begin{tabular}{lcc}
\toprule
\textbf{Method} & \textbf{MimicGen}~\cite{mandlekar2023mimicgen} & \textbf{\ours} \\
\midrule
\textbf{Demo Gen SR}$\uparrow$ & 0.11 & \textbf{0.63} \\
\bottomrule
\end{tabular}
}
\caption{Comparison of data generation success rate. \ours{} enables significantly more reliable trajectory synthesis with unseen objects as compared to Mimicgen~\cite{mandlekar2023mimicgen} that only considers object-level geometric correspondence.}
\label{tab:success_rate_comparison}
\end{table}

\subsection{Ablations}
\label{sec:exp-ablation}

\begin{figure}[h!]
    \centering
    \includegraphics[trim={1cm 1cm 1cm 1cm},clip,width=0.7\linewidth]{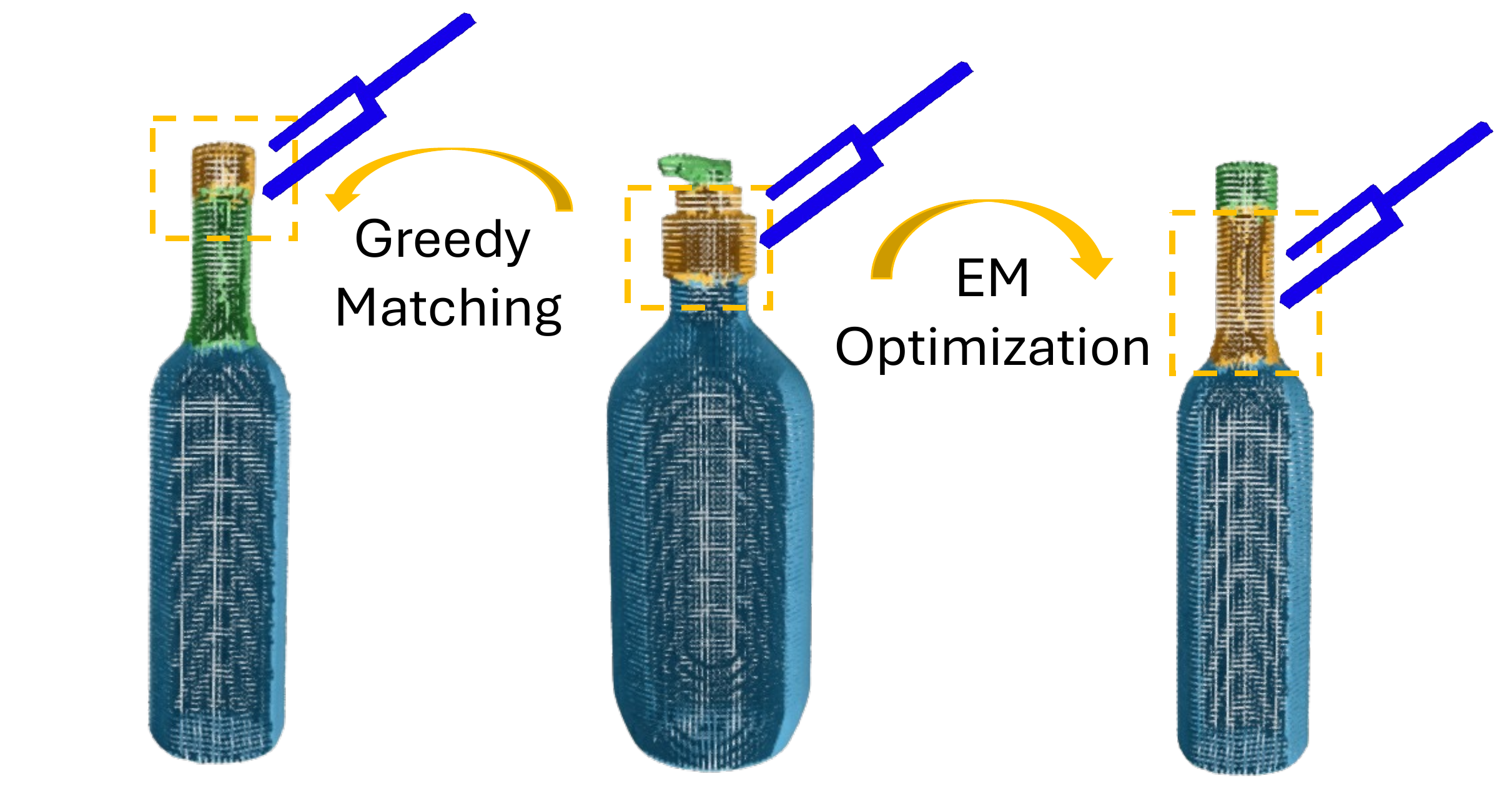}

    \caption{EM optimization yields more reliable functional alignment than greedy matching. Greedy matching aligns the source neck to the target cap due to local feature similarity, producing an unstable grasp. EM optimization updates node masses using global structural cues, correctly identifying the neck as the functional root and enabling a stable grasp.}
    \label{abla:no-em}
\end{figure}

We perform ablation studies to analyze the contributions of different components in our framework in \autoref{tab:ablation}. Removing the graph structure (``W/O GRAPH") causes the largest performance drop, reducing SR and NSS, which demonstrates that part-level structural relations are critical for accurate correspondence. In contrast, replacing DINO with CLIP (``W/O DINO, W/ CLIP'') yields performance nearly identical to the full model in SR and DTM, but with a slight decline in NSS.  Removing BFS-based saliency weighting (``W/O BFS SALIENCY") also degrades performance, highlighting the importance of propagating functional relevance from the root part. Finally, using a greedy root annotation for target feature graph without EM optimization (``ROOT ANCHOR") leads to suboptimal alignment. In this baseline, the target root is chosen greedily as the node with the closest node distance to the source interactable part. Such greedy selection ignores global structural consistency, often anchoring the matching to a locally similar but functionally suboptimal part~(refer~\autoref{abla:no-em}), reducing both SR and NSS. Overall, these results confirm that structural reasoning, appropriate feature representation, and targeted root selection collectively drive the superior correspondence quality of our method.

% \begin{table}[t]
% \centering
% \begin{tabular}{lccc}
% \toprule
% \textbf{Methods} & \textbf{SR}$\uparrow$ & \textbf{NSS}$\uparrow$ & \textbf{DTM}$\downarrow$ \\
% \midrule
% W/O DINO      & \textbf{0.79} & 0.61 & 0.06 \\
% W/O GRAPH     & 0.32 & -0.37 & 0.13 \\
% W/O DINO, W/ DIFT  & 0.68 & \textbf{0.86} & 0.04\\
% \midrule
% W/O BFS SALIENCY  & 0.63 & 0.47 & 0.04 \\
% ROOT ANCHOR  & 0.68 & 0.51 & 0.05 \\
% \midrule
% \textbf{\ours} & 0.74 & 0.61 & \textbf{0.03} \\
% % \textbf{XXX(Ours)} & 0.76 & 1.62 & 0.01 \\
% %    
% \bottomrule
% \end{tabular}
% \caption{Ablations}
% \label{tab:ablation}
% \end{table}
\begin{table}[t]
\centering
\resizebox{0.8\linewidth}{!}{
\begin{tabular}{lccc}
\toprule
\textbf{Methods} & \textbf{SR}$\uparrow$ & \textbf{NSS}$\uparrow$ & \textbf{DTM}$\downarrow$ \\
\midrule
W/O DINO      & 0.85 & 0.78 & \textbf{0.01} \\
W/O GRAPH     & 0.38 & 0.21 & 0.09 \\
W/O DINO, W/ CLIP  & 0.85 & 0.77 & \textbf{0.01}\\
\midrule
W/O BFS SALIENCY   & 0.77 & 0.71 & 0.02 \\
ROOT ANCHOR  & 0.77 & 0.65 & 0.02 \\
\midrule
\textbf{\ours} & \textbf{0.85} & \textbf{0.82} & 0.02 \\
% \textbf{XXX(Ours)} & 0.76 & 1.62 & 0.01 \\
%    
\bottomrule
\end{tabular}
}
\caption{Ablation study on key components of our correspondence framework. 
We evaluate the effect of part-level features (DINO/CLIP), graph structure, BFS-based saliency weighting, 
and EM-refined root selection.
Removing graph structure quantification significantly degrades performance, 
highlighting the importance of structural reasoning.}
\label{tab:ablation}
\end{table}

\section{Conclusion and Limitations}
\label{sec:conclusion}

% We introduced \ours, a geometry-aware correspondence framework for generalizable robotic manipulation without additional training or semantic priors.  
% By representing objects as part-based graphs and measuring structural similarity via the Unbalanced Fused Gromov–Wasserstein distance, our method establishes fine-grained geometric and functional correspondences across disjoint object categories.  
% This structure-driven approach enables both zero-shot transfer and scalable demonstration synthesis.

% Experiments show that \ours{} substantially improves zero-shot manipulation success on unseen objects and, when integrated with MimicGen, generates diverse, physically valid demonstrations that train robust manipulation policies.  
% These results highlight correspondence-driven reasoning as a promising foundation for data-efficient and open-world robotic learning.  
% Future work will extend this framework to dynamic and articulated interactions, unifying geometric understanding and data generation for lifelong manipulation.

We introduced \ours, a geometry-aware correspondence framework for generalizable robotic manipulation without extra training or semantic priors.
By representing objects as part-based graphs and comparing their structure via the Unbalanced Fused Gromov–Wasserstein~(UFGW) distance, the method recovers fine-grained geometric and functional correspondences across disjoint object categories.
This structure-centric formulation supports both zero-shot transfer and scalable demonstration generation.
Experiments show that \ours{} markedly improves zero-shot manipulation success on unseen objects and, when paired with MimicGen, produces diverse and physically valid demonstrations that train robust policies.
These findings position correspondence-driven reasoning as a strong basis for data-efficient, open-world manipulation. 

In the supplementary material, we further provide results with automated segmentation and preliminary results on articulated objects. Despite these advances, several limitations remain. Deformable objects may violate the structural assumptions of our graph matcher, leading to weaker correspondence, and our evaluation is limited to pick-and-place tasks. Extending to long-horizon, multi-contact interactions remains an important direction for future work.

\bibliographystyle{ieeenat_fullname}
\bibliography{main}


\end{document}